\documentclass{bmvc2k}
\usepackage{amsfonts}
\usepackage{enumitem}

\pdfoutput=1

\title{Multi-Stain Self-Attention Graph Multiple Instance Learning Pipeline for Histopathology Whole Slide Images}

\addauthor {Amaya Gallagher-Syed}{a.r.syed@qmul.ac.uk}{1, 2}
\addauthor{Luca Rossi}{luca.rossi@polyu.edu.hk}{3}
\addauthor{Felice Rivellese}{f.rivellese@qmul.ac.uk}{2}
\addauthor{Costantino Pitzalis}{c.pitzalis@qmul.ac.uk}{2}
\addauthor{Myles Lewis}{myles.lewis@qmul.ac.uk}{2}
\addauthor{Michael Barnes}{m.r.barnes@qmul.ac.uk}{1,2}
\addauthor{Gregory Slabaugh}{g.slabaugh@qmul.ac.uk}{2}

\addinstitution{
 Digital Environment Research Institute\\
 Queen Mary University of London\\
 London, UK
}
\addinstitution{
 William Harvey Research Institute\\
 Queen Mary University of London\\
 London, UK
}
\addinstitution{
Department of Electronic and Information Engineering\\
The Hong Kong Polytechnic University\\
Hong Kong
}

\runninghead{Gallagher-Syed et al}{MUSTANG}

\begin{document}

\maketitle

\begin{abstract}

Whole Slide Images (WSIs) present a challenging computer vision task due to their gigapixel size and presence of numerous artefacts. Yet they are a valuable resource for patient diagnosis and stratification, often representing the gold standard for diagnostic tasks. Real-world clinical datasets tend to come as sets of heterogeneous WSIs with labels present at the patient-level, with poor to no annotations. Weakly supervised attention-based multiple instance learning approaches have been developed in recent years to address these challenges, but can fail to resolve both long and short-range dependencies. Here we propose an end-to-end multi-stain self-attention graph (MUSTANG) multiple instance learning pipeline, which is designed to solve a weakly-supervised gigapixel multi-image classification task, where the label is assigned at the patient-level, but no slide-level labels or region annotations are available. The pipeline uses a self-attention based approach by restricting the operations to a highly sparse $k$-Nearest Neighbour Graph of embedded WSI patches based on the Euclidean distance. We show this approach achieves a state-of-the-art F1-score/AUC of $0.89$/$0.92$, outperforming the widely used CLAM model~\cite{Lu2021}. Our approach is highly modular and can easily be modified to suit different clinical datasets, as it only requires a patient-level label without annotations and accepts WSI sets of different sizes, as the graphs can be of varying sizes and structures. The source code can be found at \texttt{https://github.com/AmayaGS/MUSTANG}.

\end{abstract}

\section{Introduction}
\label{sec:intro}

In recent years, deep learning techniques have become the preferred methodology for analysing medical images, particularly in histology image classification \cite{AhmedtAristizabal2022}. Histopathology slide analysis is a time-consuming task that requires a qualified histopathologist. However, the digitisation of histopathology slides into Whole Slide Images (WSIs) has made it possible to automate analysis using deep learning techniques for routine workflows like subtyping, grading, and localising Regions of Interest (ROI). This automation accelerates analysis, enhances inter-observer concordance by offering a consistent benchmark, and can help overcome limitations due to restricted access to qualified professionals \cite{Lucchesi2020}. \\

WSIs are large gigapixel multi-resolution image files, which present several challenges such as their large size, high memory requirements, and heterogeneity of artefacts (variations in staining intensity, scanner used, pen marks, etc). Since WSIs can be as large as 100k $\times$ 100k pixels, relevant information may be localised in small regions of the image (such as micro-tumours or Ectopic Lymphoid Structures (ELS)), but can also depend on interactions between far away parts (macro-tissue architecture) \cite{Chen2021}. Labels are often only present at the slide or patient-level (represented by a set of WSIs), which can make it very challenging to develop classification models which accurately capture both micro and macro information.

\section{Related Work}

\paragraph{Multiple Instance Learning.} Given the gigapixel size and heterogeneity of WSIs, they present a challenging computer vision task, with many successful deep learning methods rendered computationally intractable. Most approaches use variations on the weakly supervised Multiple Instance Learning (MIL) algorithm, where the gigapixel image is divided into a set of smaller patches (e.g. $224 \times 224$ pixels), inheriting noisy slide/patient labels. Patches are then embedded into a feature vector and classified at the slide/patient level using some form of non-trainable global (e.g. max or mean) pooling on the set of instances \cite{Chen2022}. This weakly supervised learning approach has yielded clinical grade performance, despite low instance level accuracy \cite{Ilse18,Coudray2018,GhaffariLaleh2022}. Furthermore, these approaches fail at capturing long-range dependencies and in learning which regions are most relevant to the final classification. 

\paragraph{Attention.} Recent methods have introduced linear attention pooling layers \cite{Ilse18} and clustering-constrained attention pooling \cite{Lu2021}, which replaces non-trainable global pooling by a trainable weighted average aggregation layer where weights are given by a two-layered neural network \cite{Ilse18}. This approach has demonstrated high performance on publicly available datasets and crucially provides information on key instances, permitting heatmap visualisation of the attention weights associated with each image patch \cite{Lu2021}. 

\paragraph{Self-attention.} In \cite{Ilse18} and \cite{Lu2021} the attention pooling layer measures the importance of a patch embedding given the whole sequence of patch embeddings. In contrast, self-attention measures the importance of a given instance compared to all other pairwise instances, which can provide a better understanding of long-range dependencies \cite{Vaswani2017}. However, applying self-attention to the full sequence of embedded patches is computationally infeasible as the complexity grows quadratically to sequence length $O(N^2)$, rendering both runtime and memory usage problematic \cite{Wang2020}. Several approaches have been developed to improve the computational complexity of self-attention, such as relying on restricting the number of self-attention operations by inducing sparsity or by reformulating the problem via matrix factorisation for example \cite{Vaswani2017,Child2019,Wang2020,Kitaev2020}. Sparse attention methods reduce the complexity by only considering a subset of the computations in the $N \times N$ self-attention matrix $P$. In matrix factorisation approaches, $P$ is assumed to be of low rank, meaning not all elements of the matrix are linearly independent of each other \cite{Wang2020}. Intuitively, the idea that WSIs patches are not independent of each other is reasonable, suggesting that restricting self-attention operations via induced sparsity could achieve good results. The question then becomes how to restrict operations to subsets of relevant patches, while incurring minimal information loss. For example in \cite{Kitaev2020} finding the nearest neighbours of tokens in high-dimensional space is achieved via locality-sensitive hashing. 

\paragraph{Graph Neural Networks.} Graph Neural Networks (GNNs) are capable of learning hierarchical representations of graphs by propagating node features through a series of message-passing and aggregation operations. Given a graph over a set of nodes $V$, during the $k$-th message-passing iteration, the embedding $\mathbf{h}_u^{(k)}$ corresponding to each node $u \in V$ is updated according to information aggregated from the neighbours of $u$, i.e.

\begin{equation}
\begin{aligned}
\mathbf{h}_u^{(k+1)} & =\operatorname{UPDATE}^{(k)}\left(\mathbf{h}_u^{(k)}, \operatorname{AGGREGATE}^{(k)}\left(\left\{\mathbf{h}_v^{(k)}, \forall v \in \mathcal{N}(u)\right\}\right)\right) \\
& =\operatorname{UPDATE}^{(k)}\left(\mathbf{h}_u^{(k)}, \mathbf{m}_{\mathcal{N}(u)}^{(k)}\right) \,,
\end{aligned}
\end{equation}
where the neighbourhood $\mathcal{N}(u)$ is defined as the set of nodes that share an edge with $u$, UPDATE and AGGREGATE are arbitrary differentiable functions, and $\mathbf{m}_{\mathcal{N}(u)}$ is the ``message'' that is aggregated from $\mathcal{N}(u)$. At each iteration, the AGGREGATE function takes as input the set of embeddings of the nodes in $\mathcal{N}(u)$ \cite{Hamilton2020}. 

\paragraph{Graphs in Histopathology.} Although Convolutional Neural Networks (CNNs) have shown impressive performance in histopathology analysis, they are less able to capture complex neighbourhood information as they analyse local areas determined by the size of the convolutional kernel \cite{AhmedtAristizabal2022}. GNNs can better exploit these irregular relationships by preserving neighbouring information and are thus ideally suited to representing relational information despite a graph structure not being explicitly present in the data \cite{AhmedtAristizabal2022,Li2022,han2022,senior2023graph}. As presented in \cite{AhmedtAristizabal2022}, applications in histopathology can be divided into cell, patch or tissue-level graphs, with both node and graph classification approaches being employed. In particular, the MIL problem can be reformulated as a patch-graph with graph-level classification, where GNN layers are combined with pooling and readout layers to produce an end-to-end framework. Patch-graphs can be constructed using features extracted from a WSI or a set of WSIs, and then connected via edges. This has been done by connecting selected or spatially adjacent patches, as well as ``super-pixels'' \cite{Achanta2012,Adnan2020,Zheng2022,Zheng2022a}. However, in real-world settings, information across multiple WSIs is integrated by the pathologist for the purpose of disease diagnosis and subtyping \cite{Dwivedi2022}. Thus, combining the relevant information from multiple images more accurately reproduces the pathology pipeline. This only complexifies the MIL problem, as the labels are at the patient and not the slide level. Furthermore, large bodies of WSIs coming from clinical trials have few to no annotations, yet still present a valuable source of information. Histopathology deep learning frameworks therefore need to extend to real-world datasets with little curation and very noisy labels. 

\subsection{Contributions}

\begin{itemize}

\item Here we propose a novel end-to-end MUlti-STain self-AtteNtion Graph (MUSTANG) multiple instance learning pipeline. MUSTANG solves a weakly-supervised gigapixel multi-image classification task, where the label is assigned at the patient-level (across the multiple images), but no slide-level labels or region annotations are available. 

\item The pipeline introduces a self-attention based approach on multiple full gigapixel WSIs by restricting the attention operations to a highly sparse $k$-Nearest Neighbour Graph ($k$-NNG) of embedded WSI patches based on Euclidean distance.

\item Importantly, our approach does not require registration of WSIs, preprocessing or labelling of ROIs, nor any feature engineering for the embedded feature vectors, making it straightforward and flexible to apply to real-world clinical datasets.

\end{itemize}

\section{Methods}

\subsection{Dataset}

\paragraph{Rheumatoid Arthritis.}

\begin{figure}[!htbp]
\centering
\vspace*{-0.1in}
\includegraphics[width=0.48\textwidth]{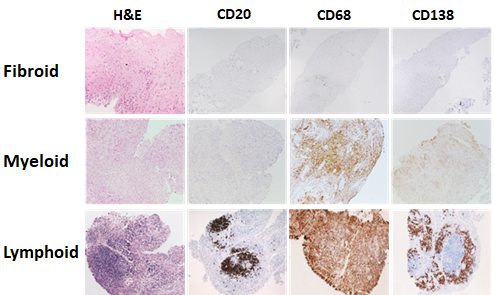}
\caption{Rheumatoid Arthritis inflammatory pathotypes Fibroid, Myeloid \& Lymphoid based on semi-quantitative analysis of synovial tissue biopsies stained with H\&E, CD20+ B cells, CD68+ macrophages and IHC+ CD138 plasma cells \cite{Humby2021}.}
\label{stain_types}
\end{figure}

MUSTANG is designed for the Rheumatoid Arthritis R4RA clinical trial dataset \cite{Rivellese2022,Humby2021}. 20 European centres recruited a total of 164 patients who underwent ultrasound-guided synovial biopsy of a clinically active joint. The synovial tissue samples were then stained with Hematoxylin \& Eosin (H\&E) and Immunochemistry (IHC) stains. H\&E provides information on overall tissue architecture and composition, while IHC identifies specific immune cells, such as CD20+ B cells, CD68+ macrophages and CD138+ plasma cells \cite{Humby2021}. Each dye contains complimentary information about the underlying disease process. Pathologists then semi-quantitatively assigned patients to one of three groups: Fibroid, Myeloid, and Lymphoid, each corresponding to a disease subtype linked to drug response and patient trajectory \cite{Dennis2014,Lewis2019}. The different stain types and disease pathotypes can be seen in Figure \ref{stain_types}. For the purpose of this study, the pathotypes Fibroid and Myeloid are aggregated and compared to the Lymphoid pathotype, as the latter more substantially differs in disease presentation, trajectory, and drug response. Samples were scanned into WSIs with .ndpi format with Hamamatsu digital scanners under 20x objectives. The dataset has a total of 651 WSIs, with a variable number of WSIs per patient.

\subsection{MUSTANG}

\begin{figure}[!t]
\centering
\includegraphics[width=\textwidth]{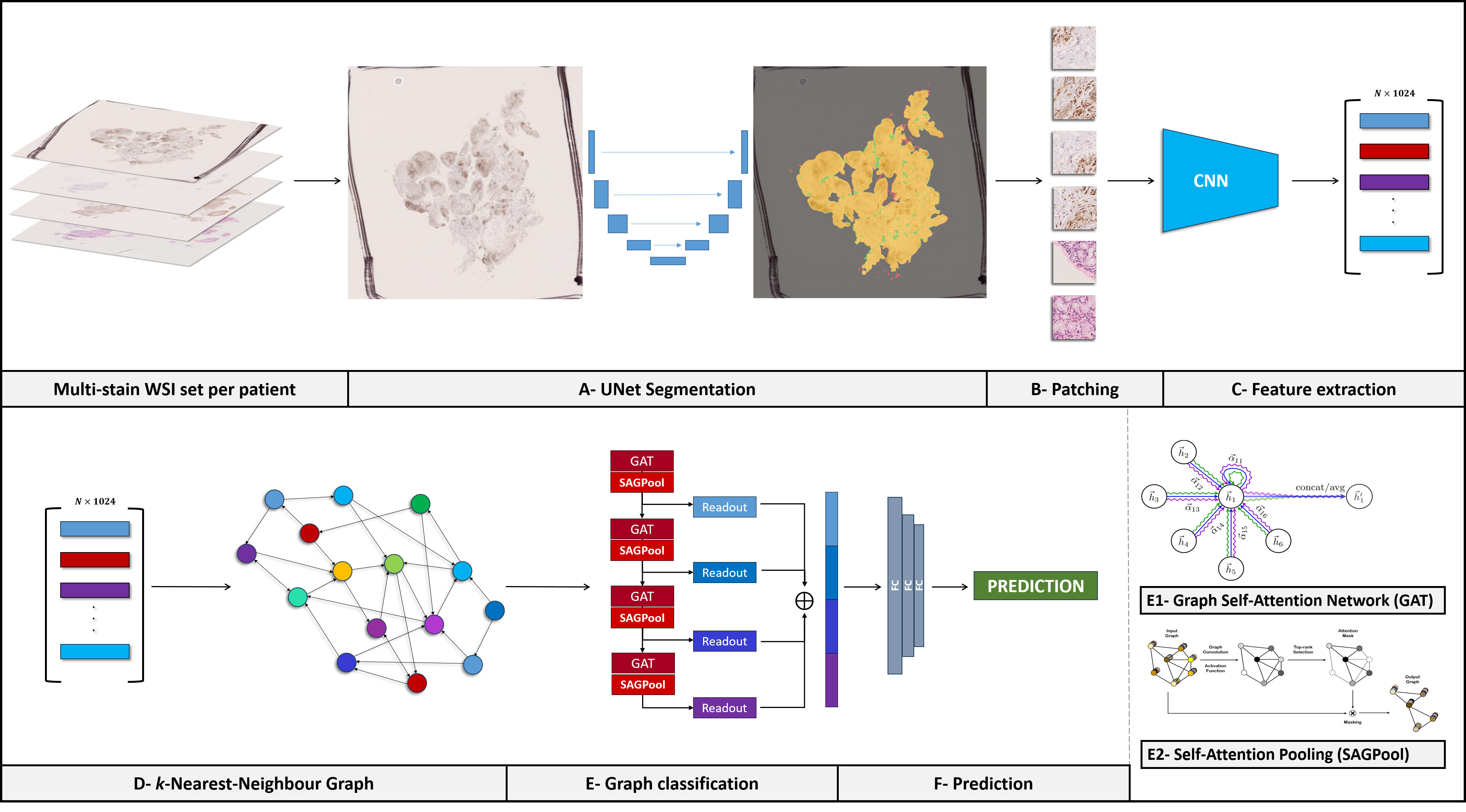}
\caption{MUSTANG pipeline. Schemas in E1 and E2 are reproduced from \cite{sagpool, Velickovic2018}}
\label{model}
\end{figure}

The MUSTANG pipeline, which is graphically represented in Figure \ref{model} is composed of: 

\medskip

\begin{itemize}[noitemsep, nolistsep]

\item \textbf{A - Segmentation}: A automated segmentation step, where UNet is used to segment tissue areas on the WSIs. The user can use the trained weights provided on our GitHub repository or use their own. 

\item \textbf{B - Patching}: After segmentation, the tissue area is divided into patches at a size chosen by the user, which can be overlapping or non-overlapping. 

\item \textbf{C - Feature extraction}: Each image patch is passed through a VGG16 CNN feature extractor and embedded into a [$1 \times 1024$] feature vector. All feature vectors from a given patient are aggregated into a matrix. The number of rows in the matrix will vary as each patient has a variable set of WSIs, each with their own dimensions. 

\item \textbf{D - $k$-Nearest-Neighbour Graph}: The matrix of feature vectors of each patient is used to create a sparse directed $k$-NNG using the Euclidean distance metric, with a default of $k=5$. The attribute of each node corresponds to a [$1 \times 1024$] feature vector. This graph is used as input to the GNN. 

\item \textbf{E - Graph classification}: The $k$-NNG is successively passed through four Graph Attention Network layers (GAT) \cite{Velickovic2018} and SAGPooling layers \cite{sagpool}. The SAGPooling readouts from each layer are concatenated and passed through three MLP layers and finally classified. 

\item \textbf{F - Prediction}: A pathotype or diagnosis prediction is obtained at the patient-level. 

\end{itemize}

\paragraph{Segmentation and patch extraction.}

From the 651 WSIs, a total 309,248 non-overlapping $224 \times 224$ pixel patches were extracted at 10x magnification from the tissue area segmented by UNet \cite{GallagherSyed2023}. The 10x magnification was chosen based on our domain knowledge of RA as a compromise to show both the macro/micro-architecture of the tissue and to reduce the number of patches for storage and computation \cite{GallagherSyed2023}. 

\paragraph{$k$-Nearest Neighbour Graph.}

The $k$-NNG is a directed graph (digraph) in which node $p$ is connected by a directed edge to node $q$, if $q$ is among the $k$ closest nodes to $p$. Here we measure the distance between two nodes in terms of the Euclidean distance between the corresponding embedded feature vectors. Notice that not all connections are reciprocal because if node $p$ is $q$'s closest neighbour, it does not follow that $q$ is also $p$'s closest neighbour. A digraph is weakly connected if there exists a path between every pair of nodes in the underlying undirected graph.


\paragraph{Graph Attention Network.}

Graph Attention Networks (GATs), based on the self-attention mechanism \cite{Bahdanau2016,Vaswani2017}, incorporate masked self-attention layers into graph convolutions and use attention weights to define a weighted sum of the neighbours:

\begin{equation}
\mathbf{m}_{\mathcal{N}(u)}=\sum_{v \in \mathcal{N}(u)} \alpha_{u, v} \mathbf{h}_v \,,
\end{equation}
where $\alpha_{u, v}$ denotes the attention on neighbour $v \in \mathcal{N}(u)$ when aggregating information at node $u$. In the original GAT paper, the attention weights are defined as:

\begin{equation}
\alpha_{u, v}=\frac{\exp \left( \mathsf{LeakyReLU}\left( \mathbf{a}^{\top}\left[\mathbf{W h}_u \oplus \mathbf{W} \mathbf{h}_v\right]\right) \right)}{\sum_{v^{\prime} \in \mathcal{N}(u)} \exp \left( \mathsf{LeakyReLU} \left(\mathbf{a}^{\top}\left[\mathbf{W} \mathbf{h}_u \oplus \mathbf{W} \mathbf{h}_{v^{\prime}}\right]\right) \right)},
\end{equation}

where $\mathbf{a}$ is a trainable attention vector, $\mathbf{W}$ is a trainable matrix and $\oplus$ denotes the concatenation operation. To stabilise the learning process multi-head attention can be used, where $m$ different attention heads are applied to compute mutually independent features in parallel, and their features are then averaged.

\paragraph{Graph Attention Pooling.}

Graph pooling is used to downsample a graph, reducing its size while seeking to preserve topological information. Global pooling methods use max or mean aggregation layers to pool all the representations of nodes in each layer, which enables graphs of different structures to be processed but tends to lose topological information. Similarly, hierarchical methods such as DiffPool \cite{Ying2019} or gPool \cite{Gao2019a}, which use a learnable vector to calculate projection scores and select the top-ranked nodes, do not fully take into account graph topology \cite{Gao2019,sagpool,Cangea2018}. SAGPool \cite{sagpool} uses the GCN defined in \cite{Kipf2017} to calculate the self-attention scores $Z \in \mathbb{R}^{N \times 1}$ as follows:

\begin{equation}
Z=\sigma\left(\tilde{D}^{-\frac{1}{2}} \tilde{A} \tilde{D}^{-\frac{1}{2}} X \Theta_{a t t}\right)\,,
\end{equation}
where $\sigma$ is the activation function, $\tilde{A} \in \mathbb{R}^{N \times N}$ is the adjacency matrix with self-connections,  $\tilde{D} \in \mathbb{R}^{N \times N}$ is the degree matrix of $\tilde{A}$, $X \in \mathbb{R}^{N \times F}$ is the matrix of input features of the graph with $N$ nodes and $F$-dimensional features, and $\Theta_{a t t} \in \mathbb{R}^{F \times 1}$ is the only parameter of the SAGPool layer \cite{sagpool}. By utilizing graph convolution to obtain self-attention scores, the result of the pooling is based on both graph and topological features. The node selection method follows \cite{Gao2019,Cangea2018, Knyazev2019} by retaining a portion of nodes of the input graph even when graphs of varying sizes and structures are input. The pooling ratio $k \in(0,1]$ is a hyperparameter that determines the number of nodes to keep. The top $\lceil k N\rceil$ nodes are selected based on the value of $Z$, i.e.

\begin{equation}
\mathrm{idx}=\operatorname{top}-\operatorname{rank}(Z,\lceil k N\rceil), \quad Z_{\text {mask }}=Z_{\mathrm{idx}}\,,
\end{equation}
where $\operatorname{top}-\operatorname{rank}$ is the function that returns the indices of the top $\lceil k N\rceil$ values, $idx$ is an indexing operation, and $Z_{m a s k}$ is the feature attention mask. Finally, the readout layer, adopted from \cite{Cangea2018}, aggregates node features to make a fixed size representation. The summarized output feature of the readout layer is

\begin{equation}
s=\frac{1}{N} \sum_{i=1}^N x_i \oplus \max _{i=1}^N x_i \,,
\end{equation}
where $N$ is the number of nodes, $x_i$ id the feature vector of the $i$-th node, and $\oplus$ denotes concatenation \cite{Cangea2018}. 

\paragraph{Benchmarking and ablation studies.}

We benchmark our method against CLAM \cite{Lu2021}, which is a clustering-constrained gated attention based MIL method, widely used within the histopathology community. We modify the framework slightly to accommodate the patient WSI set's DataLoader and use VGG16 instead of ResNet50 for feature extraction. We also try several different GNN architectures using combinations of GAT or GCN, with SAGPooling or TopKPooling (GCN\_SAG, CGN\_TopK, GAT\_SAG), to assess the importance of each component. Finally, we look at F1-scores for different values of $k$ in the $k$-NNG. 

\paragraph{Training schedule.}

This is a binary classification task with weakly-supervised labels. To obtain our results we use a 70/30 train/test split and train with the Adam optimizer $\beta_1=0.9$, $\beta_2=0.98$ and $\epsilon=10^{-9}$. We use the default parameters recommended for CLAM \cite{Lu2021}: $lr_{CLAM}= 0.0001$, with no dropout. We train MUSTANG with $lr_{MUSTANG}= 0.0001$, \textit{pooling ratio}=$0.8$, \textit{attention heads}=$2$. We train each method for 50 epochs and keep the best F1-score, whilst checking the loss is stable (i.e. that we are not underfitting). The training was prototyped locally on a commercial workstation with NVidia GPU RTX3080 and trained with QMUL's Apocrita HPC facility on an NVidia A100 GPU supported by QMUL Research-IT \cite{King2017}.

\section{Experimental Results}

\begin{figure}[h!]
\centering
\includegraphics[width=0.49\textwidth]{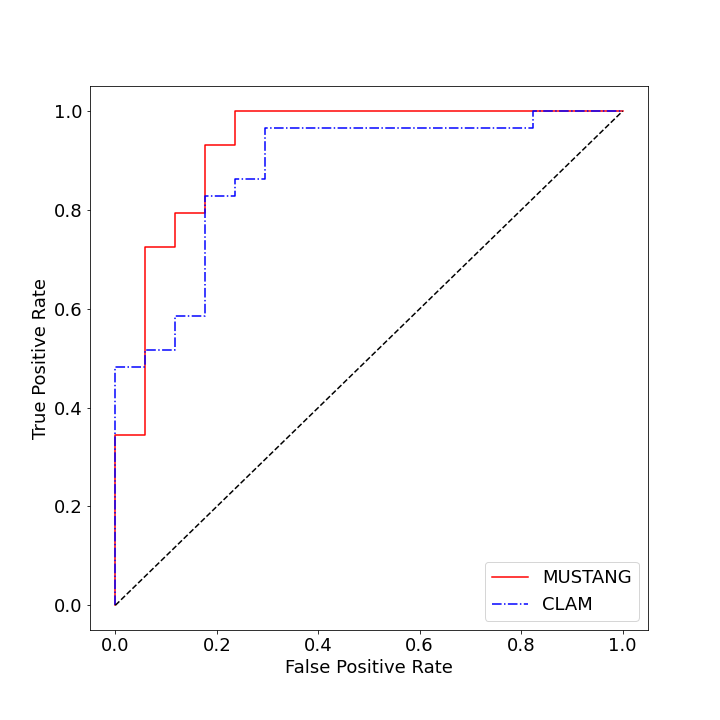}
\includegraphics[width=0.49\textwidth]{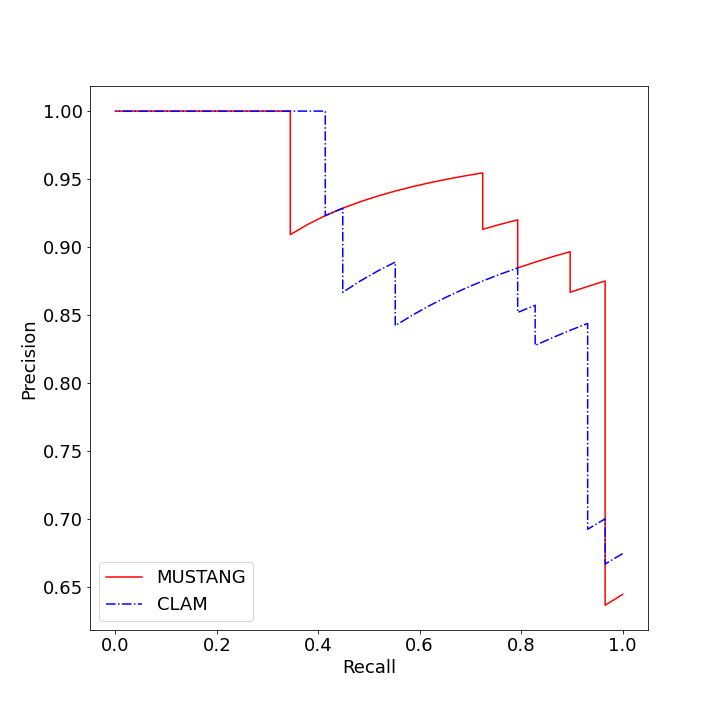}
\caption{Area Under the ROC and Precision-Recall curves for both MUSTANG (full line) and CLAM (dotted line).}
\label{prc}
\end{figure}

\paragraph{Multi-stain WSIs.} In Table \ref{multi-stain} we present the results obtained by both MUSTANG and CLAM on the multi-stain R4RA Rheumatoid Arthritis test set. MUSTANG outperforms the benchmark by $5$ percentage points for both F1-score and AUC and runtime on inference is substantially similar to that of CLAM, despite having a larger number of total parameters and relying on self-attention operations. In Figure \ref{prc}, we show MUSTANG performs well at identifying both correct and true positives (Sensitivity=$0.93$), an important consideration for healthcare.

\begin{table}[h!]
\begin{center}	
\label{tab:my-table}
\begin{tabular}{rccccccc}
\hline
\multicolumn{1}{c}{} &  F1-score & AUC & Sens & Spec & Params [M] & Test runtime [min]		\\ \hline
CLAM    & 0.84  & 0.88 & 0.86 & 0.82 &   0.47 & 10   \\ 
MUSTANG (ours)  & \textbf{0.89}  & \textbf{0.92} & \textbf{0.93} & 0.82 & 3.29 & 11 \\ \hline
\end{tabular}
\end{center}
\caption{Multi-stain F1-score, AUC, Sensitivity (Sens) and Specificity (Spec) results, total parameter number, and test runtime results.}
\label{multi-stain}
\end{table}

\begin{table}[h!]
\begin{center}	
\label{tab:my-table}
\begin{tabular}{rcccc}
\hline
\multicolumn{1}{c}{} & \multicolumn{4}{c}{Single Stain}   \\ \hline
                            & CD138   & CD68   & CD20  & HE                 \\
CLAM                        &  0.85   & 0.87  &  \textbf{0.88}  &  0.76   \\
MUSTANG (ours)                    & \textbf{0.89}    & \textbf{0.89}   & 0.87  & \textbf{0.78} \\ \hline
\end{tabular}
\end{center}
\caption{Single-stain F1-score results}
\label{single-stain}
\end{table}

\paragraph{Single-stain WSIs.} To understand if our model is able to outperform single-stain F1-score, we propagate the patient WSIs set label to each slide and rerun the method on the single-stain. The results are presented in Table \ref{single-stain}. MUSTANG outperforms CLAM, except for CD20+ staining, and obtains similar accuracy results to the multi-stain problem with 89\% F1-score for CD138 and CD68. H\&E is the worst performing stain in both cases, suggesting any IHC staining performed in the clinic would be beneficial for patient pathotype assignment. CLAM performs better on the single-stain problem than on the multi-stain one, confirming it does not fully capture long-range dependencies as the patient matrix size increases. MUSTANG does not show increased performance compared to the single-stain problem, but successfully integrates information across stains, identifying complex relations and spatial arrangements pertaining to disease subtyping. This is valuable because we do not know a priori which stains contain the most information about the disease pathotype, as most clinical datasets have not received curation. 

\paragraph{GNN model ablation.}  In Table \ref{components} we check which component parts of the GNN model provide the most accuracy gain. From the high accuracy of the GAT/GCN + SAGPool models, it is clear that the self-attention pooling topology preserving method is crucial in aggregating graph topological information which preserves the relations between nodes. 

\begin{table}[h!]
\begin{center}
\begin{tabular}{ccclc}
\hline
GNN Model   & \multicolumn{1}{r}{GAT + SAGPool} & \multicolumn{1}{r}{GAT + TopK} & GCN + SAGPool            & \multicolumn{1}{r}{GCN + TopK} \\ \hline
Multi Stain & \textbf{0.89}    & 0.67                           & \multicolumn{1}{c}{\textbf{0.82}} & 0.76                           \\ \hline
\end{tabular}
\end{center}
\caption{GNN model ablation F1-score results}
\label{components}
\end{table}

\begin{figure}[h!]
\centering
	\includegraphics[width=0.8\textwidth]{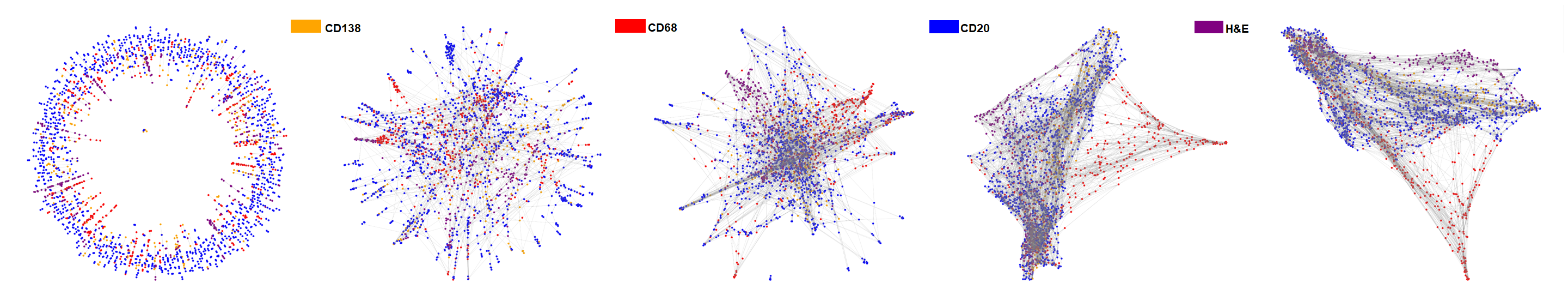}
	\caption{$k$-NNG layout for different values of $k$. From left to right, $k=1$, $k=2$, $k=3$, $k=4$ and $k=5$.}
	\label{K layout}
\end{figure}

\begin{figure}[h!]
\centering
	\includegraphics[width=0.8\textwidth]{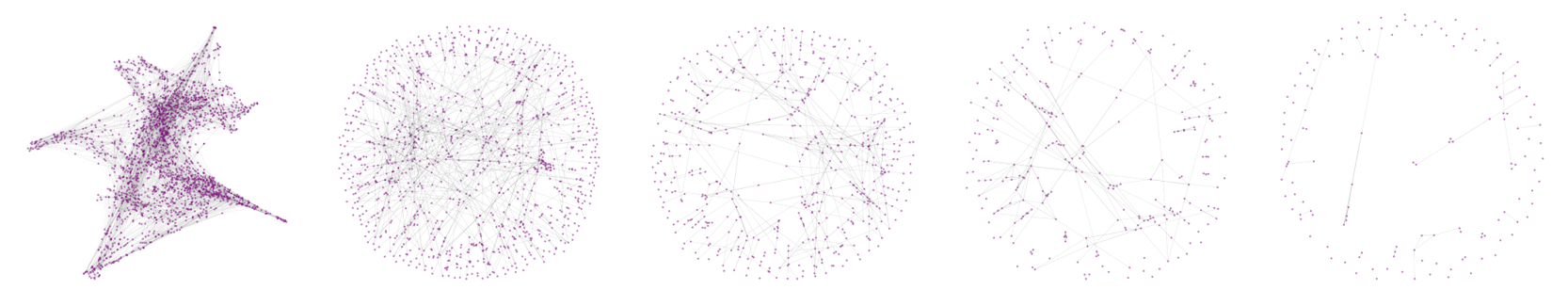}
	\caption{$k$-NNG graph connectivity layout after each GAT + SAGPool layer, with pooling ratio of 0.5}
	\label{graph-linkage}
\end{figure}

\paragraph{$k$-NNG ablation on $k$.} In Figure \ref{K layout} we show the $k$-NNG layout for different values of $k$. Up to $k=4$, the graph is not a weakly connected digraph, meaning that it has isolated nodes and subgraphs. For message-passing purposes, we reason that the initial graph should be weakly-connected in order to obtain better results. We check this assumption running MUSTANG for several values of $k$ and present the results in Table \ref{K ablation}. The $k$-NNG with $k=5$ obtains the highest score, with $k=1$ the lowest. $k=2$ already shows good accuracy, evidencing even highly sparse poorly connected graphs suffice to extract relevant signals. Higher values of $k$ also obtain good results, but a denser graphs comes at the cost of higher memory requirements and the potential to over-smooth the signal, suggesting that using a minimally weakly connected graph is a good strategy. In Figure \ref{graph-linkage} we show how the $k$-NNG graph connectivity structure changes after each GAT + SAGPool layer in the GNN model: after each layer the graph loses structure, restricting message passing to increasingly many small subgraphs. We posit the early layers aggregate macro-tissue topological information while the later layers concentrate on micro-tissue information, suggesting that graph disaggregation is not a phenomenon to avoid.  

\begin{table}[h!]
\centering
\begin{tabular}{|l|c|c|c|c|c|c|c|c|c|}
\hline
$k$        & 1 & 2 & 3 & 4 & 5 & 10 & 20 & 50 & 100 \\ \hline
Accuracy & 0.71  & 0.85 & 0.87 & 0.87 & \textbf{0.89}  & 0.83  & 0.87 & 0.85 & 0.87   \\ \hline
\end{tabular}
\caption{$k$-NNG model $k$ ablation F1-score results}
\label{K ablation}
\end{table}

\begin{figure}[h!]
\centering
\includegraphics[width=0.39\textwidth]{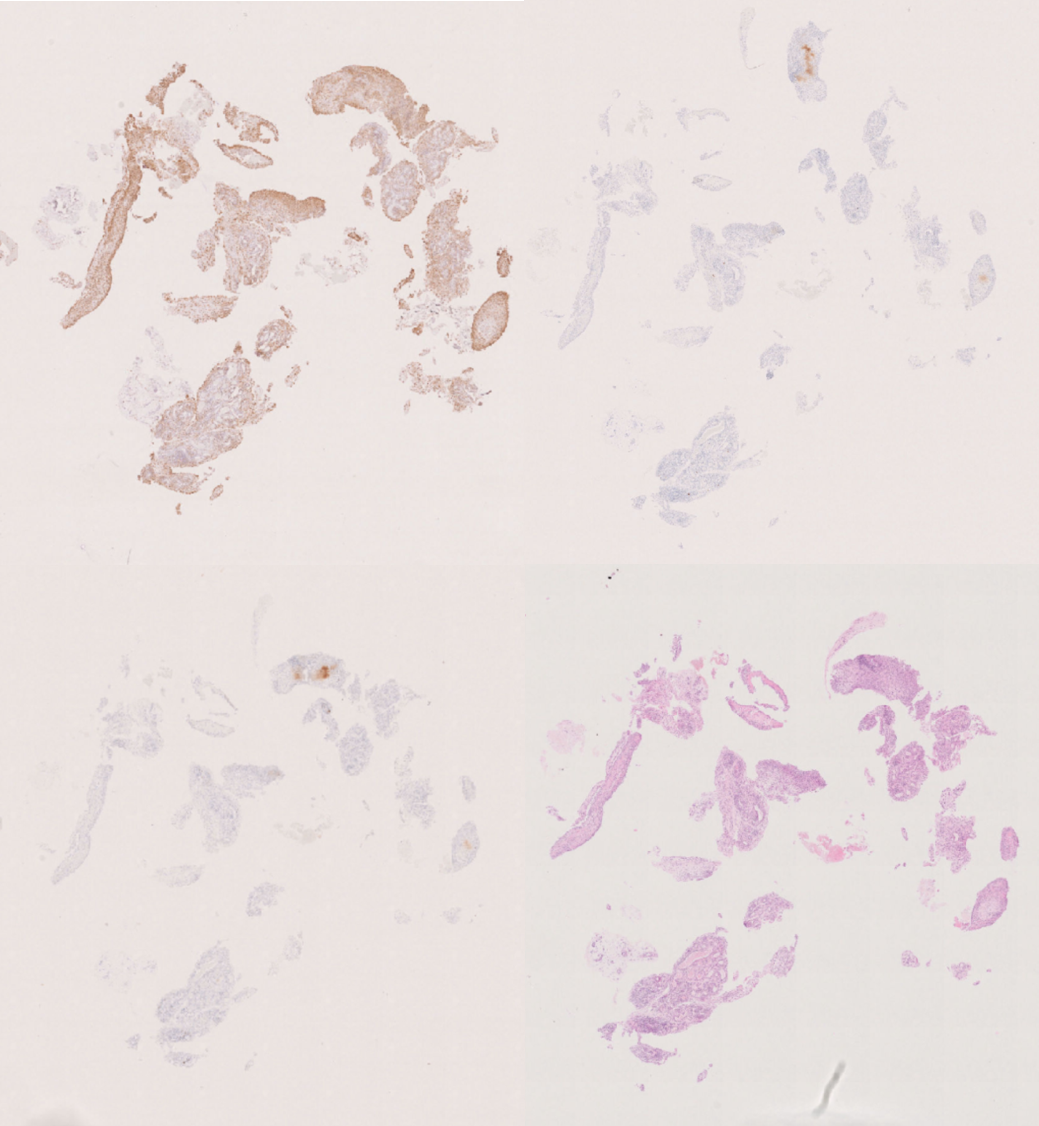}
\includegraphics[width=0.45\textwidth]{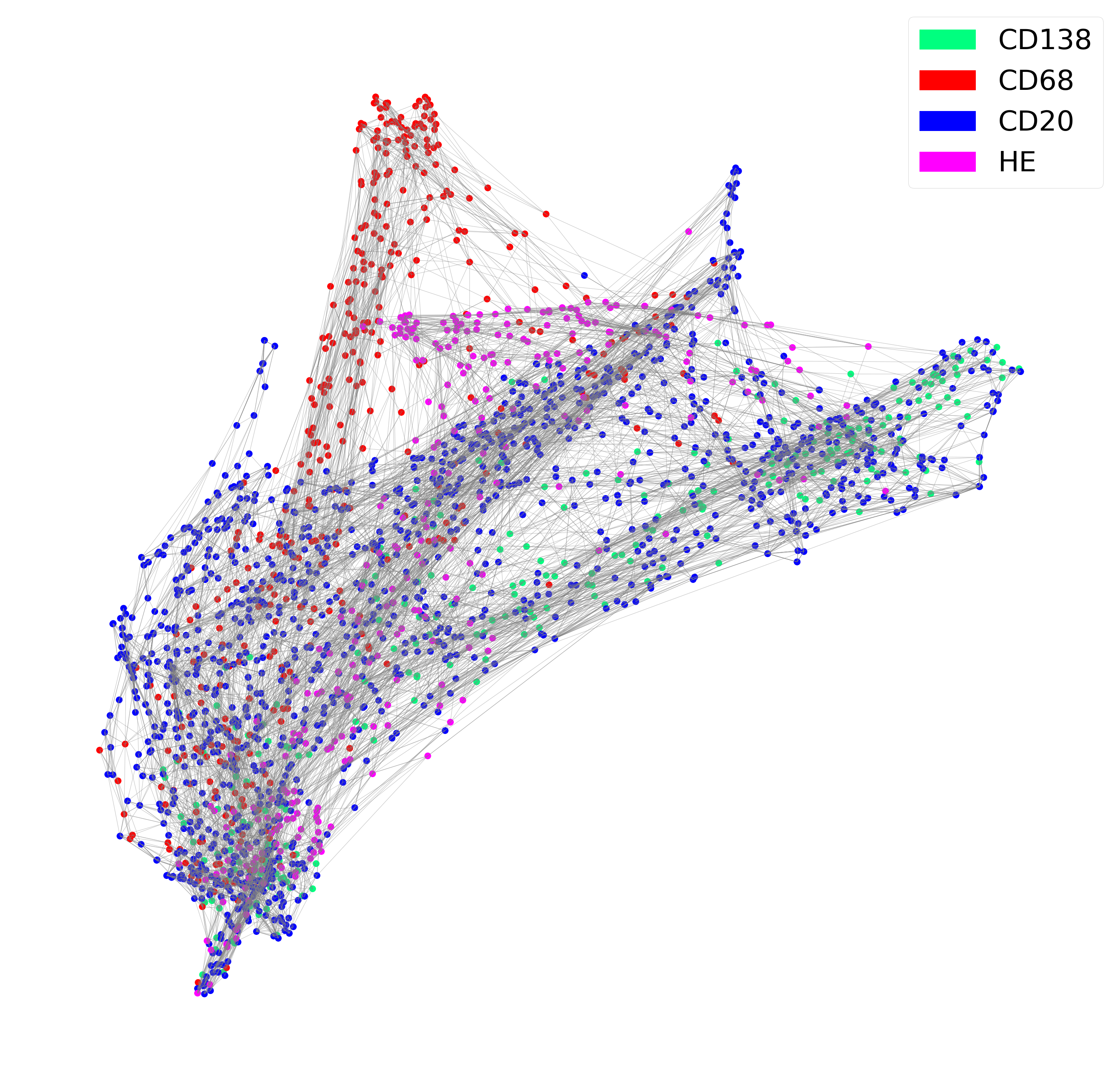}
\caption{On left a set of WSI corresponding to a patient and on right its corresponding $k$-NNG ($k=5$) plotted using Networkx's spring-layout.}
\label{connectivity}
\end{figure}

\paragraph{Graph connectivity.} We empirically check graph connectivity to see what each node is connecting to in Figure \ref{connectivity}. On the left, we show a set of WSIs for a single patient and on the right the corresponding $k$-NNG ($k=5$) graph structure plotted using Networkx spring-layout \cite{SciPyProceedings}, with nodes coloured in function of their provenance. The force-directed spring layout uses the Fruchterman-Reingold algorithm, where edges act as ``springs'' and nodes repel each other, hence resulting in closely connected nodes clustering together \cite{Fruchterman1991}. As expected nodes tend to connect to other nodes in the same WSI, but there is a good degree of mixing between WSIs, indicating information can flow between them. 

\paragraph{Limitations.} Ideally we would apply MUSTANG to publicly available datasets for comparison purposes. However, we are not aware of a public dataset with multi-stain WSI sets labelled at the patient-level, despite this being a common histopathology pipeline. We address this limitation by comparing our method against CLAM, which has strong performance on a variety of public datasets. Another limitation is that we have not yet developed a visualisation of heatmap overlays indicating the regions of interest that the model attends to. Visualisation is a crucial step for the translation of these tools into clinical practice and which remains as future work.
\vspace*{-0.1in}
\paragraph{Supplementary material.} Our supplementary material has additional results and analysis.
\section{Conclusion}

We introduced a novel end-to-end MUlti-STain self-AtteNtion Graph Multiple Instance Learning pipeline, which we call MUSTANG. The pipeline employs a self-attention based approach on multiple full gigapixel Whole Slide Images by restricting the attention operations to a highly sparse $k$-NNG of embedded WSI patches based on Euclidean distance. We show this approach achieves state-of-the-art accuracy, outperforming the widely used CLAM model. This shows that when data is highly correlated, applying self-attention operations on a very sparse matrix is sufficient to integrate both long-range dependencies and local behaviour. This approach is highly modular and can easily be modified to suit different clinical datasets, as it only requires a patient-level label without any WSI annotations and accepts WSI sets of different sizes, as the graphs can be of varying sizes and structures. 

\section*{Acknowledgements} This work was supported with funding from the Wellcome Trust (grant no. 218584/Z/19/Z). The R4RA trial was funded by the Efficacy and Mechanism Evaluation (EME) Programme, a partnership between the Medical Research Council (MRC) and the National Institute for Health and Care Research (NIHR) (grant no. 11/100/76). This work acknowledges the support of the National Institute for Health Research Barts Biomedical Research Centre (NIHR203330). 

\bibliography{MUSTANG.bib}

\begin{thebibliography}{38}
\providecommand{\natexlab}[1]{#1}
\providecommand{\url}[1]{\texttt{#1}}
\expandafter\ifx\csname urlstyle\endcsname\relax
  \providecommand{\doi}[1]{doi: #1}\else
  \providecommand{\doi}{doi: \begingroup \urlstyle{rm}\Url}\fi

\bibitem[Achanta et~al.(2012)Achanta, Shaji, Smith, Lucchi, Fua, and
  Süsstrunk]{Achanta2012}
Radhakrishna Achanta, Appu Shaji, Kevin Smith, Aurelien Lucchi, Pascal Fua, and
  Sabine Süsstrunk.
\newblock {SLIC} {Superpixels} {Compared} to {State}-of-the-{Art} {Superpixel}
  {Methods}.
\newblock \emph{IEEE Transactions on Pattern Analysis and Machine
  Intelligence}, 34\penalty0 (11):\penalty0 2274--2282, November 2012.
\newblock ISSN 1939-3539.
\newblock \doi{10.1109/TPAMI.2012.120}.

\bibitem[Adnan et~al.(2020)Adnan, Kalra, and Tizhoosh]{Adnan2020}
Mohammed Adnan, Shivam Kalra, and Hamid~R. Tizhoosh.
\newblock Representation {Learning} of {Histopathology} {Images} using {Graph}
  {Neural} {Networks}.
\newblock In \emph{2020 {IEEE}/{CVF} {Conference} on {Computer} {Vision} and
  {Pattern} {Recognition} {Workshops} ({CVPRW})}, pages 4254--4261, June 2020.
\newblock \doi{10.1109/CVPRW50498.2020.00502}.
\newblock ISSN: 2160-7516.

\bibitem[Ahmedt-Aristizabal et~al.(2022)Ahmedt-Aristizabal, Armin, Denman,
  Fookes, and Petersson]{AhmedtAristizabal2022}
David Ahmedt-Aristizabal, Mohammad~Ali Armin, Simon Denman, Clinton Fookes, and
  Lars Petersson.
\newblock A survey on graph-based deep learning for computational
  histopathology.
\newblock \emph{Computerized Medical Imaging and Graphics}, 95:\penalty0
  102027, January 2022.
\newblock ISSN 0895-6111.
\newblock \doi{10.1016/j.compmedimag.2021.102027}.
\newblock URL
  \url{https://www.sciencedirect.com/science/article/pii/S0895611121001762}.

\bibitem[Bahdanau et~al.(2016)Bahdanau, Cho, and Bengio]{Bahdanau2016}
Dzmitry Bahdanau, Kyunghyun Cho, and Yoshua Bengio.
\newblock Neural {Machine} {Translation} by {Jointly} {Learning} to {Align} and
  {Translate}.
\newblock Technical report, May 2016.
\newblock URL \url{http://arxiv.org/abs/1409.0473}.
\newblock arXiv:1409.0473 [cs, stat] type: article.

\bibitem[Cangea et~al.(2018)Cangea, Veličković, Jovanović, Kipf, and
  Liò]{Cangea2018}
Cătălina Cangea, Petar Veličković, Nikola Jovanović, Thomas Kipf, and
  Pietro Liò.
\newblock Towards {Sparse} {Hierarchical} {Graph} {Classifiers}.
\newblock Technical report, November 2018.
\newblock URL \url{http://arxiv.org/abs/1811.01287}.
\newblock arXiv:1811.01287 [cs, stat] type: article.

\bibitem[Chen et~al.(2021)Chen, Lu, Weng, Chen, Williamson, Manz, Shady, and
  Mahmood]{Chen2021}
Richard~J. Chen, Ming~Y. Lu, Wei-Hung Weng, Tiffany~Y. Chen, Drew~F.K.
  Williamson, Trevor Manz, Maha Shady, and Faisal Mahmood.
\newblock Multimodal co-attention transformer for survival prediction in
  gigapixel whole slide images.
\newblock In \emph{Proceedings of the IEEE/CVF International Conference on
  Computer Vision (ICCV)}, pages 4015--4025, October 2021.

\bibitem[Chen et~al.(2022)Chen, Chen, Li, Chen, Trister, Krishnan, and
  Mahmood]{Chen2022}
Richard~J. Chen, Chengkuan Chen, Yicong Li, Tiffany~Y. Chen, Andrew~D. Trister,
  Rahul~G. Krishnan, and Faisal Mahmood.
\newblock Scaling {Vision} {Transformers} to {Gigapixel} {Images} via
  {Hierarchical} {Self}-{Supervised} {Learning}.
\newblock In \emph{2022 {IEEE}/{CVF} {Conference} on {Computer} {Vision} and
  {Pattern} {Recognition} ({CVPR})}, pages 16123--16134, June 2022.
\newblock \doi{10.1109/CVPR52688.2022.01567}.
\newblock ISSN: 2575-7075.

\bibitem[Child et~al.(2019)Child, Gray, Radford, and Sutskever]{Child2019}
Rewon Child, Scott Gray, Alec Radford, and Ilya Sutskever.
\newblock Generating {Long} {Sequences} with {Sparse} {Transformers}.
\newblock Technical report, April 2019.
\newblock URL \url{http://arxiv.org/abs/1904.10509}.
\newblock arXiv:1904.10509 [cs, stat] type: article.

\bibitem[Coudray et~al.(2018)Coudray, Ocampo, Sakellaropoulos, Narula, Snuderl,
  Fenyö, Moreira, Razavian, and Tsirigos]{Coudray2018}
Nicolas Coudray, Paolo~Santiago Ocampo, Theodore Sakellaropoulos, Navneet
  Narula, Matija Snuderl, David Fenyö, Andre~L. Moreira, Narges Razavian, and
  Aristotelis Tsirigos.
\newblock Classification and mutation prediction from non–small cell lung
  cancer histopathology images using deep learning.
\newblock \emph{Nature Medicine}, 24\penalty0 (10):\penalty0 1559--1567,
  October 2018.
\newblock ISSN 1546-170X.
\newblock \doi{10.1038/s41591-018-0177-5}.
\newblock URL \url{https://www.nature.com/articles/s41591-018-0177-5}.

\bibitem[Dennis et~al.(2014)Dennis, Holweg, Kummerfeld, Choy, Setiadi, Hackney,
  Haverty, Gilbert, Lin, Diehl, Fischer, Song, Musselman, Klearman, Gabay,
  Kavanaugh, Endres, Fox, Martin, and Townsend]{Dennis2014}
Glynn Dennis, Cécile~TJ Holweg, Sarah~K Kummerfeld, David~F Choy, A~Francesca
  Setiadi, Jason~A Hackney, Peter~M Haverty, Houston Gilbert, Wei~Yu Lin, Lauri
  Diehl, S~Fischer, An~Song, David Musselman, Micki Klearman, Cem Gabay, Arthur
  Kavanaugh, Judith Endres, David~A Fox, Flavius Martin, and Michael~J
  Townsend.
\newblock Synovial phenotypes in rheumatoid arthritis correlate with response
  to biologic therapeutics.
\newblock \emph{Arthritis Research \& Therapy}, 16\penalty0 (2):\penalty0 R90,
  2014.
\newblock ISSN 1478-6354.
\newblock \doi{10.1186/ar4555}.
\newblock URL \url{https://www.ncbi.nlm.nih.gov/pmc/articles/PMC4060385/}.

\bibitem[Dwivedi et~al.(2022)Dwivedi, Nofallah, Pouryahya, Iyer, Leidal, Chung,
  Watkins, Billin, Myers, Abel, and Behrooz]{Dwivedi2022}
Chaitanya Dwivedi, Shima Nofallah, Maryam Pouryahya, Janani Iyer, Kenneth
  Leidal, Chuhan Chung, Timothy Watkins, Andrew Billin, Robert Myers, John
  Abel, and Ali Behrooz.
\newblock Multi {Stain} {Graph} {Fusion} for {Multimodal} {Integration} in
  {Pathology}.
\newblock pages 1835--1845, 2022.
\newblock URL
  \url{https://openaccess.thecvf.com/content/CVPR2022W/CVMI/html/Dwivedi_Multi_Stain_Graph_Fusion_for_Multimodal_Integration_in_Pathology_CVPRW_2022_paper.html}.

\bibitem[Fruchterman and Reingold(1991)]{Fruchterman1991}
Thomas M.~J. Fruchterman and Edward~M. Reingold.
\newblock Graph drawing by force-directed placement.
\newblock \emph{Software: Practice and Experience}, 21\penalty0 (11):\penalty0
  1129--1164, 1991.
\newblock ISSN 1097-024X.
\newblock \doi{10.1002/spe.4380211102}.
\newblock URL
  \url{https://onlinelibrary.wiley.com/doi/abs/10.1002/spe.4380211102}.

\bibitem[Gallagher-Syed et~al.(2023)Gallagher-Syed, Khan, Rivellese, Pitzalis,
  Lewis, Slabaugh, and Barnes]{GallagherSyed2023}
Amaya Gallagher-Syed, Abbas Khan, Felice Rivellese, Costantino Pitzalis,
  Myles~J. Lewis, Gregory Slabaugh, and Michael~R. Barnes.
\newblock Automated segmentation of rheumatoid arthritis immunohistochemistry
  stained synovial tissue.
\newblock Technical report, September 2023.
\newblock URL \url{http://arxiv.org/abs/2309.07255}.
\newblock arXiv:2309.07255 [cs, eess, q-bio] type: article.

\bibitem[Gao and Ji(2019)]{Gao2019}
Hongyang Gao and Shuiwang Ji.
\newblock Graph {U}-{Nets}.
\newblock pages 2083--2092. PMLR, May 2019.
\newblock URL \url{https://proceedings.mlr.press/v97/gao19a.html}.

\bibitem[Gao et~al.(2019)Gao, Chen, and Ji]{Gao2019a}
Hongyang Gao, Yongjun Chen, and Shuiwang Ji.
\newblock Learning {Graph} {Pooling} and {Hybrid} {Convolutional} {Operations}
  for {Text} {Representations}.
\newblock In \emph{The {World} {Wide} {Web} {Conference}}, {WWW} '19, pages
  2743--2749, New York, NY, USA, May 2019. Association for Computing Machinery.
\newblock ISBN 9781450366748.
\newblock \doi{10.1145/3308558.3313395}.
\newblock URL \url{https://doi.org/10.1145/3308558.3313395}.

\bibitem[Ghaffari~Laleh et~al.(2022)Ghaffari~Laleh, Muti, Loeffler, Echle,
  Saldanha, Mahmood, Lu, Trautwein, Langer, Dislich, Buelow, Grabsch, Brenner,
  Chang-Claude, Alwers, Brinker, Khader, Truhn, Gaisa, Boor, Hoffmeister,
  Schulz, and Kather]{GhaffariLaleh2022}
Narmin Ghaffari~Laleh, Hannah~Sophie Muti, Chiara Maria~Lavinia Loeffler,
  Amelie Echle, Oliver~Lester Saldanha, Faisal Mahmood, Ming~Y. Lu, Christian
  Trautwein, Rupert Langer, Bastian Dislich, Roman~D. Buelow, Heike~Irmgard
  Grabsch, Hermann Brenner, Jenny Chang-Claude, Elizabeth Alwers, Titus~J.
  Brinker, Firas Khader, Daniel Truhn, Nadine~T. Gaisa, Peter Boor, Michael
  Hoffmeister, Volkmar Schulz, and Jakob~Nikolas Kather.
\newblock Benchmarking weakly-supervised deep learning pipelines for whole
  slide classification in computational pathology.
\newblock \emph{Medical Image Analysis}, 79:\penalty0 102474, July 2022.
\newblock ISSN 1361-8415.
\newblock \doi{10.1016/j.media.2022.102474}.
\newblock URL
  \url{https://www.sciencedirect.com/science/article/pii/S1361841522001219}.

\bibitem[Hagberg et~al.(2008)Hagberg, Schult, and Swart]{SciPyProceedings}
Aric~A. Hagberg, Daniel~A. Schult, and Pieter~J. Swart.
\newblock Exploring network structure, dynamics, and function using networkx.
\newblock In Ga\"el Varoquaux, Travis Vaught, and Jarrod Millman, editors,
  \emph{Proceedings of the 7th Python in Science Conference}, pages 11 -- 15,
  Pasadena, CA USA, 2008.

\bibitem[Hamilton()]{Hamilton2020}
William~L. Hamilton.
\newblock Graph representation learning.
\newblock \emph{Synthesis Lectures on Artificial Intelligence and Machine
  Learning}, 14\penalty0 (3):\penalty0 1--159.

\bibitem[Han et~al.(2022)Han, Wang, Guo, Tang, and Wu]{han2022}
Kai Han, Yunhe Wang, Jianyuan Guo, Yehui Tang, and Enhua Wu.
\newblock Vision {GNN}: An image is worth graph of nodes.
\newblock In Alice~H. Oh, Alekh Agarwal, Danielle Belgrave, and Kyunghyun Cho,
  editors, \emph{Advances in Neural Information Processing Systems}, 2022.
\newblock URL \url{https://openreview.net/forum?id=htM1WJZVB2I}.

\bibitem[Humby et~al.(2021)Humby, Durez, Buch, Lewis, Rizvi, Rivellese,
  Nerviani, Giorli, Mahto, Montecucco, Lauwerys, Ng, Ho, Bombardieri, Romão,
  Verschueren, Kelly, Sainaghi, Gendi, Dasgupta, Cauli, Reynolds, Cañete,
  Moots, Taylor, Edwards, Isaacs, Sasieni, Choy, Pitzalis, Thompson, Bugatti,
  Bellan, Congia, Holroyd, Pratt, Fonseca, White, Warren, Peel, Hands,
  Fossati-Jimack, Hadfield, Thorborn, Ramirez, and Celis]{Humby2021}
Frances Humby, Patrick Durez, Maya~H. Buch, Myles~J. Lewis, Hasan Rizvi, Felice
  Rivellese, Alessandra Nerviani, Giovanni Giorli, Arti Mahto, Carlomaurizio
  Montecucco, Bernard Lauwerys, Nora Ng, Pauline Ho, Michele Bombardieri,
  Vasco~C. Romão, Patrick Verschueren, Stephen Kelly, Pier~Paolo Sainaghi,
  Nagui Gendi, Bhaskar Dasgupta, Alberto Cauli, Piero Reynolds, Juan~D.
  Cañete, Robert Moots, Peter~C. Taylor, Christopher~J. Edwards, John Isaacs,
  Peter Sasieni, Ernest Choy, Costantino Pitzalis, Charlotte Thompson, Serena
  Bugatti, Mattia Bellan, Mattia Congia, Christopher Holroyd, Arthur Pratt,
  João Eurico Cabral~da Fonseca, Laura White, Louise Warren, Joanna Peel,
  Rebecca Hands, Liliane Fossati-Jimack, Gaye Hadfield, Georgina Thorborn,
  Julio Ramirez, and Raquel Celis.
\newblock Rituximab versus tocilizumab in anti-{TNF} inadequate responder
  patients with rheumatoid arthritis ({R4RA}): 16-week outcomes of a
  stratified, biopsy-driven, multicentre, open-label, phase 4 randomised
  controlled trial.
\newblock \emph{The Lancet}, 397\penalty0 (10271):\penalty0 305--317, January
  2021.
\newblock ISSN 0140-6736, 1474-547X.
\newblock \doi{10.1016/S0140-6736(20)32341-2}.
\newblock URL
  \url{https://www.thelancet.com/journals/lancet/article/PIIS0140-6736(20)32341-2/fulltext}.

\bibitem[Ilse et~al.(2018)Ilse, Tomczak, and Welling]{Ilse18}
M.~Ilse, J.~Tomczak, and M.~Welling.
\newblock Attention-based deep multiple instance learning.
\newblock In \emph{Proc. 35th ICML}, volume~80, pages 2127--2136, 2018.
\newblock URL \url{https://proceedings.mlr.press/v80/ilse18a.html}.

\bibitem[King et~al.(2017)King, Butcher, and Zalewski]{King2017}
Thomas King, Simon Butcher, and Lukasz Zalewski.
\newblock Apocrita - {High} {Performance} {Computing} {Cluster} for {Queen}
  {Mary} {University} of {London}.
\newblock March 2017.
\newblock URL \url{https://zenodo.org/record/438045}.

\bibitem[Kipf and Welling(2017)]{Kipf2017}
Thomas~N. Kipf and Max Welling.
\newblock Semi-{Supervised} {Classification} with {Graph} {Convolutional}
  {Networks}.
\newblock Technical report, February 2017.
\newblock URL \url{http://arxiv.org/abs/1609.02907}.
\newblock arXiv:1609.02907 [cs, stat] type: article.

\bibitem[Kitaev et~al.(2020)Kitaev, Kaiser, and Levskaya]{Kitaev2020}
Nikita Kitaev, Łukasz Kaiser, and Anselm Levskaya.
\newblock Reformer: {The} {Efficient} {Transformer}.
\newblock Technical report, February 2020.
\newblock URL \url{http://arxiv.org/abs/2001.04451}.
\newblock arXiv:2001.04451 [cs, stat] type: article.

\bibitem[Knyazev et~al.(2019)Knyazev, Taylor, and Amer]{Knyazev2019}
Boris Knyazev, Graham~W Taylor, and Mohamed Amer.
\newblock Understanding {Attention} and {Generalization} in {Graph} {Neural}
  {Networks}.
\newblock In \emph{Advances in {Neural} {Information} {Processing} {Systems}},
  volume~32. Curran Associates, Inc., 2019.
\newblock URL
  \url{https://proceedings.neurips.cc/paper_files/paper/2019/hash/4c5bcfec8584af0d967f1ab10179ca4b-Abstract.html}.

\bibitem[Lee et~al.(2019)Lee, Lee, and Kang]{sagpool}
Junhyun Lee, Inyeop Lee, and Jaewoo Kang.
\newblock Self-attention graph pooling.
\newblock In \emph{Proceedings of the 36th International Conference on Machine
  Learning}, 09--15 Jun 2019.

\bibitem[Lewis et~al.(2019)Lewis, Barnes, Blighe, Goldmann, Rana, Hackney,
  Ramamoorthi, John, Watson, Kummerfeld, Hands, Riahi, Rocher-Ros, Rivellese,
  Humby, Kelly, Bombardieri, Ng, DiCicco, van~der Heijde, Landewé, van der
  Helm-van Mil, Cauli, McInnes, Buckley, Choy, Taylor, Townsend, and
  Pitzalis]{Lewis2019}
Myles~J. Lewis, Michael~R. Barnes, Kevin Blighe, Katriona Goldmann, Sharmila
  Rana, Jason~A. Hackney, Nandhini Ramamoorthi, Christopher~R. John, David~S.
  Watson, Sarah~K. Kummerfeld, Rebecca Hands, Sudeh Riahi, Vidalba Rocher-Ros,
  Felice Rivellese, Frances Humby, Stephen Kelly, Michele Bombardieri, Nora Ng,
  Maria DiCicco, Désirée van~der Heijde, Robert Landewé, Annette van der
  Helm-van Mil, Alberto Cauli, Iain~B. McInnes, Christopher~D. Buckley, Ernest
  Choy, Peter~C. Taylor, Michael~J. Townsend, and Costantino Pitzalis.
\newblock Molecular {Portraits} of {Early} {Rheumatoid} {Arthritis} {Identify}
  {Clinical} and {Treatment} {Response} {Phenotypes}.
\newblock \emph{Cell Reports}, 28\penalty0 (9):\penalty0 2455--2470.e5, August
  2019.
\newblock ISSN 2211-1247.
\newblock \doi{10.1016/j.celrep.2019.07.091}.
\newblock URL
  \url{https://www.sciencedirect.com/science/article/pii/S2211124719310071}.

\bibitem[Li et~al.(2022)Li, Huang, and Zitnik]{Li2022}
Michelle~M. Li, Kexin Huang, and Marinka Zitnik.
\newblock Graph {Representation} {Learning} in {Biomedicine}.
\newblock Technical report, June 2022.
\newblock URL \url{http://arxiv.org/abs/2104.04883}.
\newblock arXiv:2104.04883 [cs, q-bio] type: article.

\bibitem[Lu et~al.(2021)Lu, Williamson, Chen, et~al.]{Lu2021}
M.~Y. Lu, D.~F.~K. Williamson, T.~Y. Chen, et~al.
\newblock Data-efficient and weakly supervised computational pathology on
  whole-slide images.
\newblock \emph{Nat. Biomed. Eng}, 5\penalty0 (6):\penalty0 555--570, June
  2021.
\newblock ISSN 2157-846X.
\newblock \doi{10.1038/s41551-020-00682-w}.
\newblock URL \url{https://www.nature.com/articles/s41551-020-00682-w}.

\bibitem[Lucchesi et~al.(2020)Lucchesi, Pontarini, Donati,
  et~al.]{Lucchesi2020}
D.~Lucchesi, E.~Pontarini, V.~Donati, et~al.
\newblock The use of digital image analysis in the histological assessment of
  {Sjögren}'s syndrome salivary glands improves inter-rater agreement and
  facilitates multicentre data harmonisation.
\newblock \emph{Clin. and Exp. Rheumatology}, 38 Suppl 126\penalty0
  (4):\penalty0 180--188, 2020.
\newblock ISSN 0392-856X.

\bibitem[Rivellese et~al.(2022)Rivellese, Surace, Goldmann, Sciacca, Çubuk,
  Giorli, John, Nerviani, Fossati-Jimack, Thorborn, Ahmed, Prediletto, Church,
  Hudson, Warren, McKeigue, Humby, Bombardieri, Barnes, Lewis, and
  Pitzalis]{Rivellese2022}
Felice Rivellese, Anna E.~A. Surace, Katriona Goldmann, Elisabetta Sciacca,
  Cankut Çubuk, Giovanni Giorli, Christopher~R. John, Alessandra Nerviani,
  Liliane Fossati-Jimack, Georgina Thorborn, Manzoor Ahmed, Edoardo Prediletto,
  Sarah~E. Church, Briana~M. Hudson, Sarah~E. Warren, Paul~M. McKeigue, Frances
  Humby, Michele Bombardieri, Michael~R. Barnes, Myles~J. Lewis, and Costantino
  Pitzalis.
\newblock Rituximab versus tocilizumab in rheumatoid arthritis: synovial
  biopsy-based biomarker analysis of the phase 4 {R4RA} randomized trial.
\newblock \emph{Nature Medicine}, 28\penalty0 (6):\penalty0 1256--1268, June
  2022.
\newblock ISSN 1546-170X.
\newblock \doi{10.1038/s41591-022-01789-0}.
\newblock URL \url{https://www.nature.com/articles/s41591-022-01789-0}.

\bibitem[Senior et~al.(2023)Senior, Slabaugh, Yuan, and Rossi]{senior2023graph}
Henry Senior, Gregory Slabaugh, Shanxin Yuan, and Luca Rossi.
\newblock Graph neural networks in vision-language image understanding: A
  survey.
\newblock \emph{arXiv preprint arXiv:2303.03761}, 2023.

\bibitem[Vaswani et~al.(2017)Vaswani, Shazeer, Parmar, Uszkoreit, Jones, Gomez,
  Kaiser, and Polosukhin]{Vaswani2017}
Ashish Vaswani, Noam Shazeer, Niki Parmar, Jakob Uszkoreit, Llion Jones,
  Aidan~N Gomez, Łukasz Kaiser, and Illia Polosukhin.
\newblock Attention is {All} you {Need}.
\newblock In \emph{Advances in {Neural} {Information} {Processing} {Systems}},
  volume~30. Curran Associates, Inc., 2017.
\newblock URL
  \url{https://proceedings.neurips.cc/paper/2017/hash/3f5ee243547dee91fbd053c1c4a845aa-Abstract.html}.

\bibitem[Veličković et~al.(2018)Veličković, Cucurull, Casanova, Romero,
  Liò, and Bengio]{Velickovic2018}
Petar Veličković, Guillem Cucurull, Arantxa Casanova, Adriana Romero, Pietro
  Liò, and Yoshua Bengio.
\newblock Graph {Attention} {Networks}.
\newblock Technical report, February 2018.
\newblock URL \url{http://arxiv.org/abs/1710.10903}.
\newblock arXiv:1710.10903 [cs, stat] type: article.

\bibitem[Wang et~al.(2020)Wang, Li, Khabsa, Fang, and Ma]{Wang2020}
Sinong Wang, Belinda~Z. Li, Madian Khabsa, Han Fang, and Hao Ma.
\newblock Linformer: {Self}-{Attention} with {Linear} {Complexity}.
\newblock Technical report, June 2020.
\newblock URL \url{http://arxiv.org/abs/2006.04768}.
\newblock arXiv:2006.04768 [cs, stat] version: 1 type: article.

\bibitem[Ying et~al.(2019)Ying, You, Morris, Ren, Hamilton, and
  Leskovec]{Ying2019}
Rex Ying, Jiaxuan You, Christopher Morris, Xiang Ren, William~L. Hamilton, and
  Jure Leskovec.
\newblock Hierarchical {Graph} {Representation} {Learning} with
  {Differentiable} {Pooling}.
\newblock Technical report, February 2019.
\newblock URL \url{http://arxiv.org/abs/1806.08804}.
\newblock arXiv:1806.08804 [cs, stat] type: article.

\bibitem[Zheng et~al.(2022{\natexlab{a}})Zheng, Gindra, Green, Burks, Betke,
  Beane, and Kolachalama]{Zheng2022}
Yi~Zheng, Rushin~H. Gindra, Emily~J. Green, Eric~J. Burks, Margrit Betke,
  Jennifer~E. Beane, and Vijaya~B. Kolachalama.
\newblock A {Graph}-{Transformer} for {Whole} {Slide} {Image} {Classification}.
\newblock \emph{IEEE Transactions on Medical Imaging}, 41\penalty0
  (11):\penalty0 3003--3015, November 2022{\natexlab{a}}.
\newblock ISSN 1558-254X.
\newblock \doi{10.1109/TMI.2022.3176598}.

\bibitem[Zheng et~al.(2022{\natexlab{b}})Zheng, Jiang, Shi, Xie, Zhang, Luo,
  Hu, Sun, Jiang, and Xue]{Zheng2022a}
Yushan Zheng, Zhiguo Jiang, Jun Shi, Fengying Xie, Haopeng Zhang, Wei Luo,
  Dingyi Hu, Shujiao Sun, Zhongmin Jiang, and Chenghai Xue.
\newblock Encoding histopathology whole slide images with location-aware graphs
  for diagnostically relevant regions retrieval.
\newblock \emph{Medical Image Analysis}, 76:\penalty0 102308, February
  2022{\natexlab{b}}.
\newblock ISSN 1361-8415.
\newblock \doi{10.1016/j.media.2021.102308}.
\newblock URL
  \url{https://www.sciencedirect.com/science/article/pii/S1361841521003534}.

\end{thebibliography}


\begin{thebibliography}{2}
\providecommand{\natexlab}[1]{#1}
\providecommand{\url}[1]{\texttt{#1}}
\expandafter\ifx\csname urlstyle\endcsname\relax
  \providecommand{\doi}[1]{doi: #1}\else
  \providecommand{\doi}{doi: \begingroup \urlstyle{rm}\Url}\fi

\bibitem[Cordonnier et~al.(2021)Cordonnier, Loukas, and Jaggi]{Cordonnier2021}
Jean-Baptiste Cordonnier, Andreas Loukas, and Martin Jaggi.
\newblock Multi-{Head} {Attention}: {Collaborate} {Instead} of {Concatenate}.
\newblock Technical report, May 2021.
\newblock URL \url{http://arxiv.org/abs/2006.16362}.
\newblock arXiv:2006.16362 [cs, stat] type: article.

\bibitem[Wang et~al.(2021)Wang, Wang, Yuan, Gu, and Huang]{Wang2021}
Zhaokang Wang, Yunpan Wang, Chunfeng Yuan, Rong Gu, and Yihua Huang.
\newblock Empirical analysis of performance bottlenecks in graph neural network
  training and inference with {GPUs}.
\newblock \emph{Neurocomputing}, 446:\penalty0 165--191, July 2021.
\newblock ISSN 0925-2312.
\newblock \doi{10.1016/j.neucom.2021.03.015}.
\newblock URL
  \url{https://www.sciencedirect.com/science/article/pii/S0925231221003659}.

\end{thebibliography}
\end{document}



\section*{Supplementary material}

\subsection*{Memory Usage}

\begin{figure}[h!]
\centering
\includegraphics[width=0.45\textwidth]{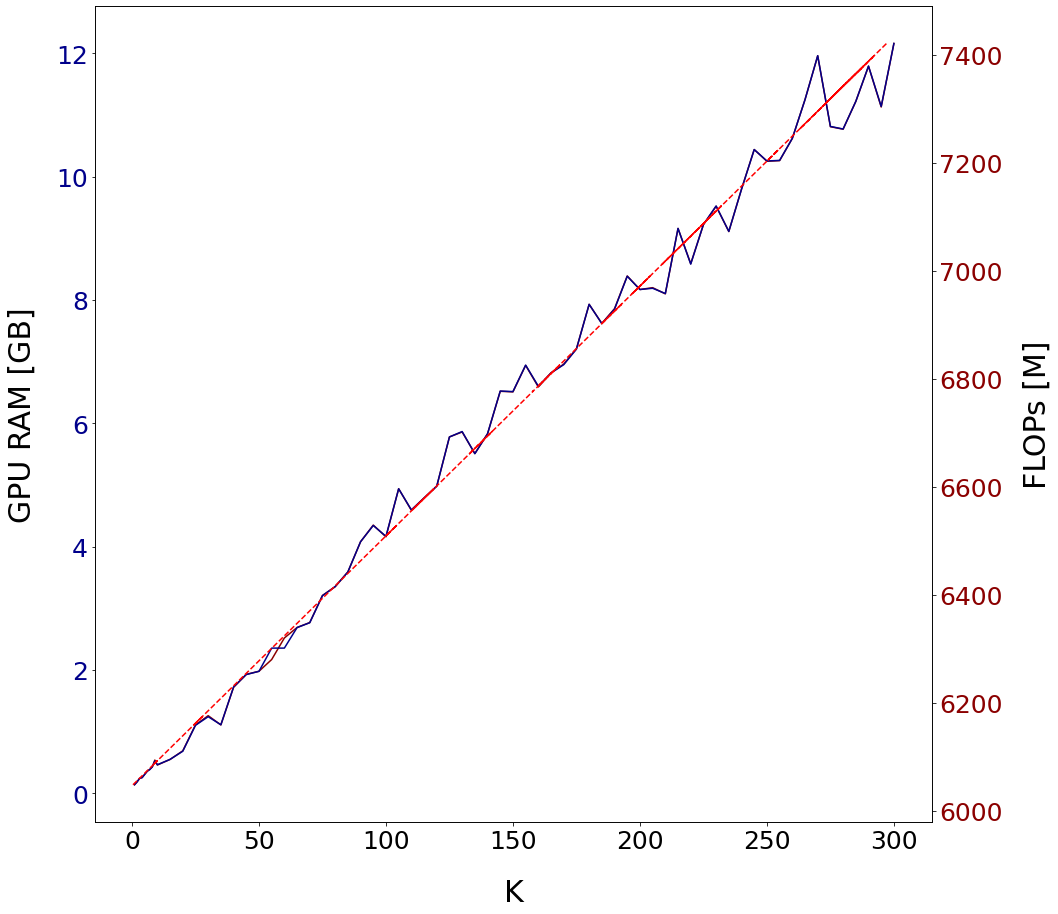}
\includegraphics[width=0.49\textwidth]{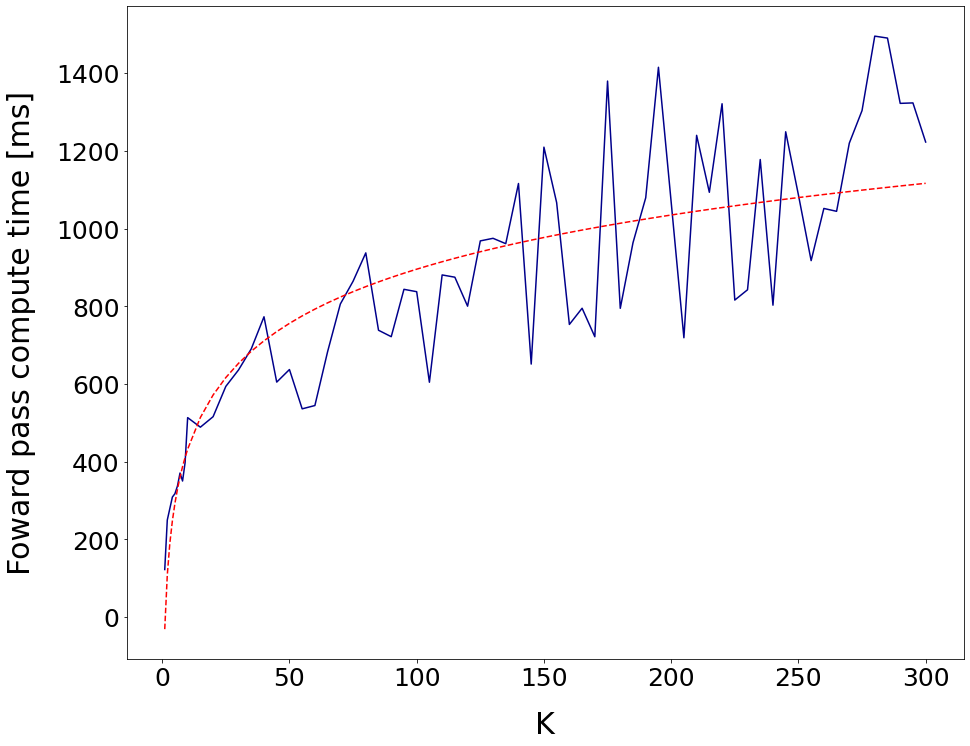}
\caption{\textit{Memory and resource usage per $k$ in a 2000 nodes $k$-NNG for a model forward pass. Left: GPU RAM [GB] usage. Right: Compute time in ms}}
\label{ram}
\end{figure}

In Figure \ref{ram} we show the increase in GPU RAM [GB] cost and FLOPs [M] in a forward pass for different values of connectivity $K$ in a $k$-NNG with $2000$ nodes, approximately the average graph size for the dataset. Despite scaling linearly in the observed range, said $k$-NNG with a $k = 100$ already requires $4$ GB RAM during the forward pass. This can quickly become prohibitive for any work done on a commercial workstation, as GPUs have limited on-chip memory \cite{Wang2021}. This highlights the need to use the minimally connected graph, so high memory needs don't prevent users from performing training and inference. This value $k$ is a parameter of the model and requires initial hyperparameter tuning to choose an optimal value.

\subsection*{Further model ablation}

We conduct further ablation on the model, to disaggregate the contribution of each part. We run the GAT\_SAGPool model with between 1 to 5 layers and we look at the number of attention heads, going from 1 to 8.  

\paragraph{Number of layers}

In Table \ref{layers} we show the increase in accuracy with each additional GAT + SAGPool layer. We note that 5 layers also obtain best F1-score, however at a higher computational cost. 

\begin{table}[h!]
\begin{center}
\begin{tabular}{rc}
\hline
 & GAT\_SAGPool \\ \hline
1 layer   & 0.80       \\
2 layer   & 0.83       \\
3 layer   & 0.87       \\
4 layer   & 0.89       \\
5 layer   & 0.89       \\ \hline
\end{tabular}
\end{center}
\caption{F1-score for a GAT\_SAGPool model with 1 to 5 layers}
\label{layers}
\end{table}

\paragraph{Number of heads}

From looking at the results in Table \ref{heads} we see 2 or 8 averaged attention heads obtain best F1-score, yet both 1 and 4 also heads obtain close results. Because this indicates the model does not need many attention heads for optimal performance, using 2 attention heads seems sufficient \cite{Cordonnier2021}, however this is a hyperparameter that can be chosen by the user based on the specifics of their dataset. 

\begin{table}[h!]
\begin{center}
\begin{tabular}{rc}
\hline
\textbf{} & F1-score  \\ \hline
1 head    &   0.87   \\
2 heads   &   0.89        \\
4 heads   &   0.87           \\
8 heads   &   0.89             \\ \hline
\end{tabular}
\end{center}
\caption{F1-score for different numbers of averaged attention heads. }
\label{heads}
\end{table}

\bibliography{MUSTANG_Supplementary.bib}